\documentclass{article}

     \usepackage[nonatbib, final]{neurips_2019}

\usepackage[utf8]{inputenc} % allow utf-8 input
\usepackage[T1]{fontenc}    % use 8-bit T1 fonts
\usepackage{hyperref}       % hyperlinks
\usepackage{url}            % simple URL typesetting
\usepackage{booktabs}       % professional-quality tables
\usepackage{amsfonts}       % blackboard math symbols
\usepackage{nicefrac}       % compact symbols for 1/2, etc.
\usepackage{microtype}      % microtypography
\usepackage{amsmath}
\usepackage{amssymb}
\usepackage{subfig}
\usepackage{graphicx}
\usepackage{ bbold }
\usepackage{algorithm}
\usepackage{algorithmic}

\DeclareMathOperator*{\argmin}{argmin}

\newcommand{\R}{{\rm I\!R}}
\title{Wasserstein Neural Processes}

\author{%
Andrew Carr \And Jared Nielsen \And David Wingate
}

\begin{document}

\maketitle

\begin{abstract}
Neural Processes (NPs) \cite{NP:follow} are a class of models that learn a mapping from a context set of input-output pairs to a distribution over functions. They are traditionally trained using maximum likelihood with a KL divergence regularization term. We show that there are desirable classes of problems where NPs, with this loss, fail to learn any reasonable distribution. We also show that this drawback is solved by using approximations of Wasserstein distance which calculates optimal transport distances even for distributions of disjoint support. We give experimental justification for our method and demonstrate performance. These Wasserstein Neural Processes (WNPs) maintain all of the benefits of traditional NPs while being able to approximate a new class of function mappings. 
\end{abstract}

\section{Introduction} \label{introduction}

Gaussian Processes (GPs) are analytically tractable, nonparametric probabilistic models that are widely used for nonlinear regression problems.  They  map an input $\vec{x_i} \in \R^{d_x}$ to an output $\vec{y_i} \in \R^{d_y}$ by defining a collection of jointly Gaussian conditional distributions. These distributions can be conditioned on an arbitrary number of context points $X_C : = (\vec{x_i})_{i \in C}$ and their associated outputs $Y_C$. Given the context points, an arbitrary number of target points $X_T : = (\vec{x_i})_{i \in T}$, with their outputs $Y_T$, can be modeled using the conditional distribution $P(X_T|X_C,Y_C)$. This modeling is invariant to the ordering of context points, and to the ordering of targets.  Formally, they define a nonparametric distribution over smooth functions, and leverage the almost magical nature of Gaussian distributions to both solve regression tasks and provide confidence intervals about the solution.  However, GPs are computationally expensive, and the strong Gaussian assumptions they make are not always appropriate.

Neural Processes (NPs) can be loosely thought of as the neural generalization of GPs.  They blend the strengths of GPs and deep neural networks: like GPs, NPs are a class of models that learn a mapping from a context set of input-output pairs to a distribution over functions.  Also like GPs, NPs can be conditioned on an arbitrary set of context points, but unlike GPs (which have quadratic complexity), NPs have linear complexity in the number of context points \cite{NP:follow}.  
NPs inherit the high model capacity of deep neural networks, which gives them the ability to fit a wide range of distributions, and can be blended with the latest deep learning best practices (for example, NPs were recently extended using attention \cite{kim2018attentive} to overcome a fundamental underfitting problem which lead to inaccurate function predictions).

Importantly, these NPs still make Gaussian assumptions; furthermore, the training algorithm of these Neural Processes uses a loss function that relies on maximum-likelihood and the KL divergence metric. Recently in the machine learning community there has been much discussion \cite{gan-zoo} around generative modeling, and the use of Wasserstein divergences. Wasserstein distance is a desirable tool because it can measure distance between two probability distributions, even if the support of those distributions is disjoint. The KL divergence, however, is undefined in such cases (or infinity); this corresponds to situations where the data likelihood is exactly zero, which can happen in the case of misspecified models.

%Therefore, replacing KL divergence with Wasserstein distance is an attractive option when distributions are disjoint. 

This begs the question: can we simultaneously improve the maximum likelihood training of NPs, and replace the Gaussian assumption in NPs with a more general model - ideally one \emph{where we do not need to make any assumptions about the data likelihood}?

More formally, it is known that in the limit of data maximizing the expected log probability of the data is equivalent to minimizing the KL divergence between the distribution implied by a model $P_\theta$ and the true data distribution $Q$, which may not be known: 
\begin{equation}
    \argmin_{\theta} D_{KL}(Q | P_{\theta}) = \argmin_{\theta} \mathop{\mathbb{E}}_Q \left [ - \log ( P_{\theta} ) \right ]
    \label{likelihood}
\end{equation}

% \begin{align}
%     D_{KL}[P(x \vert \theta^*) \, \Vert \, P(x \vert \theta)] \\
%     = \mathbb{E}_{x \sim P(x \vert \theta^*)}\left[\log \frac{P(x \vert \theta^*)}{P(x \vert \theta)} \right] \\
%         = \mathbb{E}_{x \sim P(x \vert \theta^*)}\left[\log \, P(x \vert \theta^*) - \log \, P(x \vert \theta) \right] \\
%         = \mathbb{E}_{x \sim P(x \vert \theta^*)}\left[\log \, P(x \vert \theta^*) \right] - \mathbb{E}_{x \sim P(x \vert \theta^*)}\left[\log \, P(x \vert \theta) \right]
% \end{align}

However, if the two distributions $Q$ and $P_{\theta}$ have disjoint support (i.e., if even one data point lies outside the support of the model), the KL divergence will be undefined (or infinite) due to the evaluation of $\log(0)$. Therefore, it has been shown recently \cite{arjovsky2017wassgan} that substituting Wasserstein for KL in (eq \ref{likelihood}) gives stable performance for non-overlapping distributions in the same space. This leads to using Wasserstein distance as a proxy for the KL divergence interpretation of maximum likelihood learning. 
% \begin{equation}
%     \argmin_{\theta} W_{KL}(Q | P_{\theta}) \approx \argmin_{\theta} \mathop{\mathbb{E}}_Q \left [ - \log ( P_{\theta} ) \right ]
%     \label{wasslikelihood}
% \end{equation}
This new framing is valuable in the misspecified case where data likelihood is zero under the model and additionally for the case when likelihood is unknown (such as in high dimensional data supported on low-dimensional manifolds, such as natural images), or incalculable. However, in all cases, it is required that we can draw samples from the target distribution.

The central contribution of this work is a powerful non-linear class of Wasserstein Neural Processes that can learn a variety of distributions when the data are misspecified under the model, and when the likelihood is unknown or intractable. We do so by replacing the traditional maximum likelihood loss function with a Wasserstein divergence.  We discuss computationally efficient approximations (including Sliced Wasserstein Distance) and evaluate performance in several experimental settings.

\section{Background}

\subsection{Wasserstein Distance}

As the founder of Optimal Transport theory, Gaspard Monge \cite{monge1781memoire} was interested in the problem of a worker with a shovel moving a large pile of sand. The worker's goal is to erect a target pile of sand with a prescribed shape. The worker wishes to minimize their overall effort in carrying shovelfuls of sand. Monge formulated the problem as such and worked to solve it. Over the years since then, Optimal Transport has grown into a mature and exciting field \cite{villani2008optimal}. In particular, the Wasserstein distance has seen much recent attention due to its nice theoretical properties and myriad applications \cite{peyre2018computational}. These properties enable the study of distance between two probability distributions $\mu$, $\nu$ for generative modeling.

The $p$-Wasserstein distance, $p \in [1, \infty]$, between $\mu$, $\nu$ is defined to be the solution to the original Optimal Transport problem \cite{villani2008optimal} in (eq. ~\ref{wassdist}). Where $\Gamma(\mu, \nu)$ is the joint probability distribution with marginals $\mu$, $\nu$ and $d^p(\cdot, \cdot)$ is the cost function. 
\begin{equation}
    W_p(\mu, \nu) = \left ( \inf_{\gamma \in \Gamma(\mu, \nu)} \int_{X \times Y} d^p(x,y)d\gamma(x,y) \right)^{\frac{1}{p}}
    \label{wassdist}
\end{equation}
In general, this distance is computationally intractable, in large part due to the infimum. There are, however, a number of methods used to approximate this distance. Firstly, by recalling the Kantorovich-Rubinstein duality \cite{villani2008optimal, arjovsky2017wassgan} as shown in (eq. ~\ref{duality} with $p=1$).
\begin{equation}
    W(\mu, \nu) = \sup_{||f||_L \leq 1} \mathbb{E}_{x \sim \mu}[f(x)] - \mathbb{E}_{x \sim \nu}[f(x)]
    \label{duality}
\end{equation}
It is important to note that the supremum is taken over all of the $1$-Lipschitz functions. From this form emerges an adversarial algorithm for Wasserstein distance calculation. This is explored in a number of GAN works, with \cite{gulrajani2017improved} offering good performance with the addition of a gradient penalty.
Additionally, by using a similar dual form a two-step computation that makes use of linear programming can be employed to exactly calculate the Wasserstein distance \cite{Liu2018ATC}. This method has favorable performance and is also used in generative modeling with the WGAN-TS.
Another prominent method used to calculate Wasserstein distance that is much faster to compute is that of Sinkhorn distances \cite{marko2013sinkhorn}. This method makes use of Sinkhorn-Knopp’s matrix scaling algorithm to compute the Wasserstein distance.

In the case of empirical one-dimensional distributions, the $p$-Wasserstein distance is calculated by sorting the samples from both distributions and calculating the average distance between the sorted samples (according to the cost function). This is a fast operation requiring $\mathcal{O}(n)$ operations in the best case and $\mathcal{O}(n\log n)$ in the worst case with $n$ the number of samples from each distribution \cite{kolourig2019generalized}. This simple computation is taken advantage of in the computation of the sliced Wasserstein distance.

\subsection{Sliced Wasserstein Distance}
\label{slicedwasssect}

The sliced Wasserstein distance is qualitatively similar to Wasserstein distance, but as alluded to above is much easier to compute since it only depends on computations across one-dimensional distributions \cite{kolourib2016sliced}. Intuitively, sliced Wasserstein distance is an aggregation of linear projects from high-dimensional distributions to one-dimensional distributions where regular Wasserstein distance is then calculated. The projection process is done via the Radon transform \cite{kolouri2018swauto}. The one-dimensional projects are defined in (eq. ~\ref{projectsliced}).
\begin{equation}
    \mathcal{R}p_X(t;\theta) = \int_X p_X(x)\delta(t-\theta\cdot x)dx, \quad \forall \theta \in S^{d-1}, \quad \forall t \in \R
    \label{projectsliced}
\end{equation}
Then, by using these one-dimensional projections \cite{Bon13a} the sliced Wasserstein distance is defined in (eq. ~\ref{sliced}):
\begin{equation}
    SW_p(\mu, \nu) = \left ( \int_{S^{d-1}} W_p(\mathcal{R}_{\mu}(\cdot; \theta), \mathcal{R}_{\nu}(\cdot; \theta))d\theta \right )^{\frac{1}{p}}
    \label{sliced}
\end{equation}
It is important to note that this map $SW_p$ inherits its being a distance from the fact that $W_p$ is a distance. Similarly, the two distances induce the same topology on compact sets \cite{San15a}. The integral in (eq. \ref{sliced}) is approximated by a Monte Carlo sampling variant of normal draws on $S^{d-1}$ which are averaged across samples as in (eq. \ref{approxsliced})
\begin{equation}
SW_p(\mu, \nu) \approx \left (\frac{1}{N} \sum_{i = 1}^{N} W_p(\mathcal{R}_{\mu}(\cdot; \theta_i), \mathcal{R}_{\nu}(\cdot; \theta_i))\right )^{\frac{1}{p}}
    \label{approxsliced}
\end{equation}

\subsection{Conditional Neural Processes}
First introduced in \cite{CNP:baby}, Conditional Neural Processes (CNPs) are loosely based on traditional Gaussian Processes. 
%CNPs map an input $\vec{x_i} \in \R^{d_x}$ to an output $\vec{y_i} \in \R^{d_y}$. It does this by defining a collection of conditional distributions that can be realized by conditioning on an arbitrary number of context points $X_C : = (\vec{x_i})_{i \in C}$ and their associated outputs $Y_C$. Then an arbitrary number of target points $X_T : = (\vec{x_i})_{i \in T}$, with their outputs $Y_T$ can be modeled using the conditional distribution.
Like GPs, CNPs are invariant to ordering of context points, and ordering of targets. It is important to note that, while the model is defined for arbitrary context points $C$ and target points $T$, it is common practice (which we follow) to use $ C \subset T$. 

The context point data is aggregated using a commutative operation $\oplus$ that takes elements in some $\R^d$ and maps them into a single element in the same space. In the literature, this is referred to as the $r_C$ context vector. We see that $r_C$ summarizes the information present in the observed context points.
Formally, the CNP is learning the following conditional distribution.
\begin{align}
P(Y_T | X_T, X_C, Y_C) \iff P(Y_T | X_T, r_C)
\end{align}
In practice, this is done by first passing the context points through a Neural Network $h_{\theta}$ to obtain a fixed length embedding $r_i$ of each individual context point. These context point representation vectors are aggregated with $\oplus$ to form $r_C$. The target points $X_T$ are then decoded, conditional to $r_C$, to obtain the desired output distribution $z_i$ the models the target outputs $y_i$.

More formally, this process can be defined as follows.
\begin{align}
    r_i = h_{\theta}(\vec{x_i}) && \forall \vec{x_i} \in X_C\\
    r_C = r_1 \oplus r_2 \oplus r_3 \oplus \cdots \oplus r_n \\
    z_i = g_{\phi}(\vec{y_i} | r_C) && \forall \vec{y_i} \in X_T
\end{align}

\begin{figure}
\centering
\includegraphics[scale=0.3]{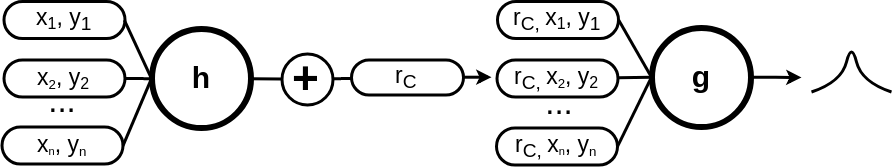}
\caption{Conditional Neural Process Architecture. Traditionally, the data is maximized under the output distribution. In this work, we use the sliced Wasserstein distance. Also note that $\oplus$ is an arbitrary commutative aggregation function and $h$, $g$ are parameterized neural networks. As an example, the context/target points could be $(x,y)$ coordinates and pixel intensities if an image is being modeled.}
\label{fig:cnp_plot}
\end{figure}

\subsection{Neural Processes}

Neural Processes are an extension of CNPs where a parameterized Normal $s_C := s(x_C, y_C)$ latent path is added to the model architecture. The purpose of this path is to model global dependencies between context and target predictions \cite{NP:follow}. The computational complexity of the Neural Processes is $\mathcal{O}(n + m)$ for $n$ contexts and $m$ targets. Additionally, \cite{NP:follow} show that in addition to being \textit{scalable} these models are \textit{flexible} and \textit{invariant} to permutations in context points. Many of these benefits are shared in with CNPs, albeit with CNPs showing slightly worse experimental performance.

With the addition of the latent path, a new loss function is required. The encoder and decoder parameters are maximized using ELBO (eq \ref{elbo}). A random subset of context points is used to predict a set of target points. With a KL regularization term that encourages the summarization of the context to be close to the summarization of the targets. 
\begin{equation}
    \log p(y_T | x_T, x_C, y_C) \geq \mathbb{E}_{q(z | s_T)} \left [ \log p(y_T | x_T, r_C, z)  \right ] - D_{KL}(q(z | s_T) || q(z|s_C))
    \label{elbo}
\end{equation}
Work has been done \cite{kim2018attentive} to explore other aggregation operations such as attention \cite{bahdanau2014neural}. These extensions solve a fundamental underfitting problems that plague both CNPs and NPs. These Attentive Neural Processes (ANP) introduce several types of attention into the model architecture in place of the neural networks which greatly improves performance at the cost of computational complexity which was shown to be $\mathcal{O}(n(n+m))$. It is important to note, however, that the introduction of attention causes the models to train more quickly (i.e., fewer iterations) which is faster in wall-clock time even if inference is slower. The ANPs utilize a fundamentally different architecture while still maximizing the ELBO loss in (eq \ref{elbo}).

\section{Wasserstein Neural Processes}

Our objective in combining Optimal Transport with Neural Processes is to model a desirable class of functions that are either misspecified, or with computationally intractable likelihood. We choose to use a CNP (fig \ref{fig:cnp_plot}) as our Neural Process architecture because CNPs are the simplest variant of the NP family with which we can illustrate the benefits of Wasserstein distance. Additionally, due to model simplicity, we can better ablate the performance and assign proper credit to the inclusion of Wasserstein distance. Extending WNPs to use a latent path, or attention, would be an interesting direction for future research.

To be precise, we use a parameterized Neural Network of shape and size determined according to task specifications. The input context $x,y$ points are encoded and aggregated, as in traditional CNPs, using a commutative operation into an $r_C$ of fixed length. The desired target $x$ points (typically the entire set of points) are then conditioned (using concatenation) and decoded to produce a synthetic set of target $y$ points. At this point, Wasserstein Neural Processes deviate from traditional NPs. Typically, as mentioned above, the target $y$ likelihood would be calculated according to the output distribution (which we do as a comparison in the experiments). However, in our case, the decoded target $y$ points are samples from an implicitly modeled distribution. As such, since the likelihood may be difficult to calculate, or non-existent, WNPs use sliced Wasserstein distance as a learning signal. This calculation is put forth in Algorithm \ref{alg:swdalgorithm}.

The sliced Wasserstein distance gives the benefit of scalable distance computation between two potentially disjoint distributions. Therefore, WNPs can model a larger variety of distributions than traditional NPs as discussed above. 

\begin{algorithm}[]
\caption{Wasserstein Neural Processes.}
\label{alg:wnpalgorithm}
\textbf{Require}: $p_0$, lower bound percentage of edges to sample as context points. $p_1$, corresponding upper bound.\\
\textbf{Require}: $\theta_0$, initial encoder parameters. $\phi_0$ initial decoder parameters. \\
\begin{algorithmic}[1] %[1] enables line numbers
\STATE Let $X$ input data in $\R^d$
\FOR{$t = 0, \cdots, n_{\text{epochs}}$ }
\FOR{$x_i$ in $X$}
\STATE Sample $p \leftarrow \text{unif}(p_0, p_1)$
\STATE Assign $n_{\text{context points}} \leftarrow p \cdot |\text{Target Points}|$
\STATE Sparsely Sample $x_i^{cp} \leftarrow x_i|_{n_{\text{context points}}}$ and associated $y_i^{cp}$
\STATE Obtain one-hot encoding of $x_i^{cp}$ points and concatenate with $y_i^{cp}$ as $F^{cp}$
\STATE Encode and aggregate $r_C \leftarrow h_{\theta}(F^{cp})$
\STATE Decode $\Tilde{x_i} \leftarrow g_{\phi}(F | r_C)$
\STATE Calculate Sliced Wasserstein Distance $l \leftarrow \mathcal{W}(\Tilde{x_i}, x_i)$
\STATE Step Optimizer
\ENDFOR
\ENDFOR
\end{algorithmic}
\end{algorithm}

\begin{algorithm}[]
\caption{Sliced Wasserstein Distance}
\label{alg:swdalgorithm}
\textbf{Require}: $n \in [5,50]$, number of desired projections. \\
\textbf{Require}: $d$, dimension of embedding space. \\
\textbf{Require}: $p$, power of desired Wasserstein distance. \\
\begin{algorithmic}[1] %[1] enables line numbers
\STATE Let $X$ be the empirical synthetic distribution and $Y$ be samples from the true distribution
\STATE Sample projections $P$ from $S^d$ i.e., $\mathcal{N}$(size$=(n, d)$) and normalize using dimension-wise $L_2$
\STATE Project $\hat{X} = X \cdot P^T$
\STATE Project $\hat{Y} = Y \cdot P^T$
\STATE Sort $\hat{X}^T$ and $\hat{Y}^T$
\STATE Return $(\hat{X}^T - \hat{Y}^T)^p$
\end{algorithmic}
\end{algorithm}

\section{Experiments}

As discussed in section \ref{introduction}, traditional Neural Processes are limited by their use of the KL divergence, both explicitly and implicitly. In this section, we demonstrate that Wasserstein Neural Processes can learn effectively under conditions that cause a traditional NP to fail. The first two experiments illustrate NPs' fundamental failing of relying on the likelihood function for learning. The final experiment shows the ability of WNPs to work on larger scale tasks. 

\subsection{Misspecified Models - Linear Regression with Uniform Noise}

In standard linear regression, parameters $\mathbf{\beta} = (\beta_1, \beta_2, \cdots, \beta_n)^T$ are estimated in the model
\begin{equation}
    \hat{Y} = \mathbf{\beta} x + b + \eta
\end{equation}
where a stochastic noise model $\eta$ is used to model disturbances. Traditionally, $\eta \sim \mathcal{N}(\mu, \sigma)$ for some parameterized Normal distribution. In such a traditional case, Neural Processes can learn the a conditional distribution over the data $x$. This is because all the data has a calculable likelihood under any setting of the parameters.

\begin{figure}
    \centering
    \subfloat{{\includegraphics[scale=0.4]{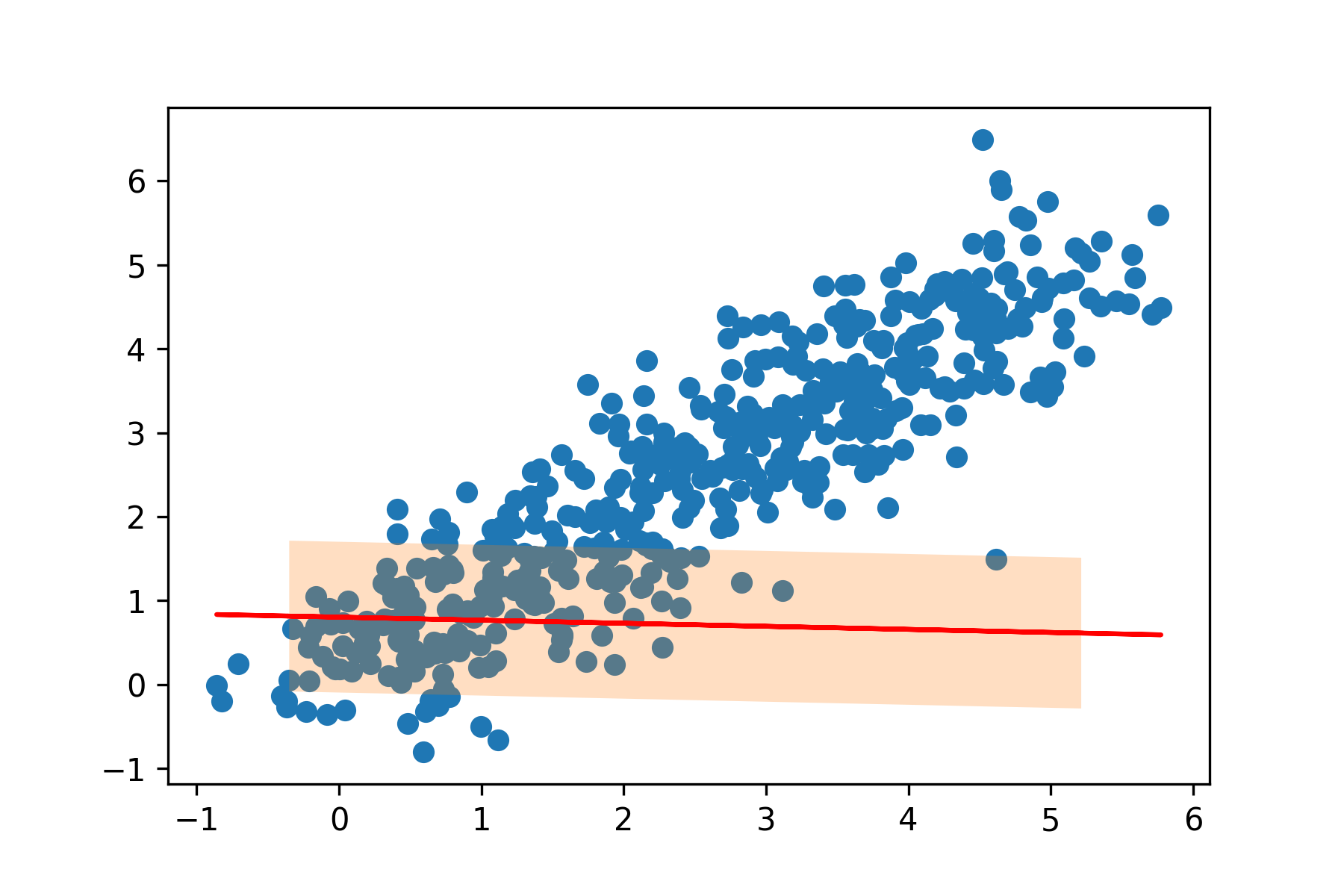} }}
    \qquad
    \subfloat{{\includegraphics[scale=0.4]{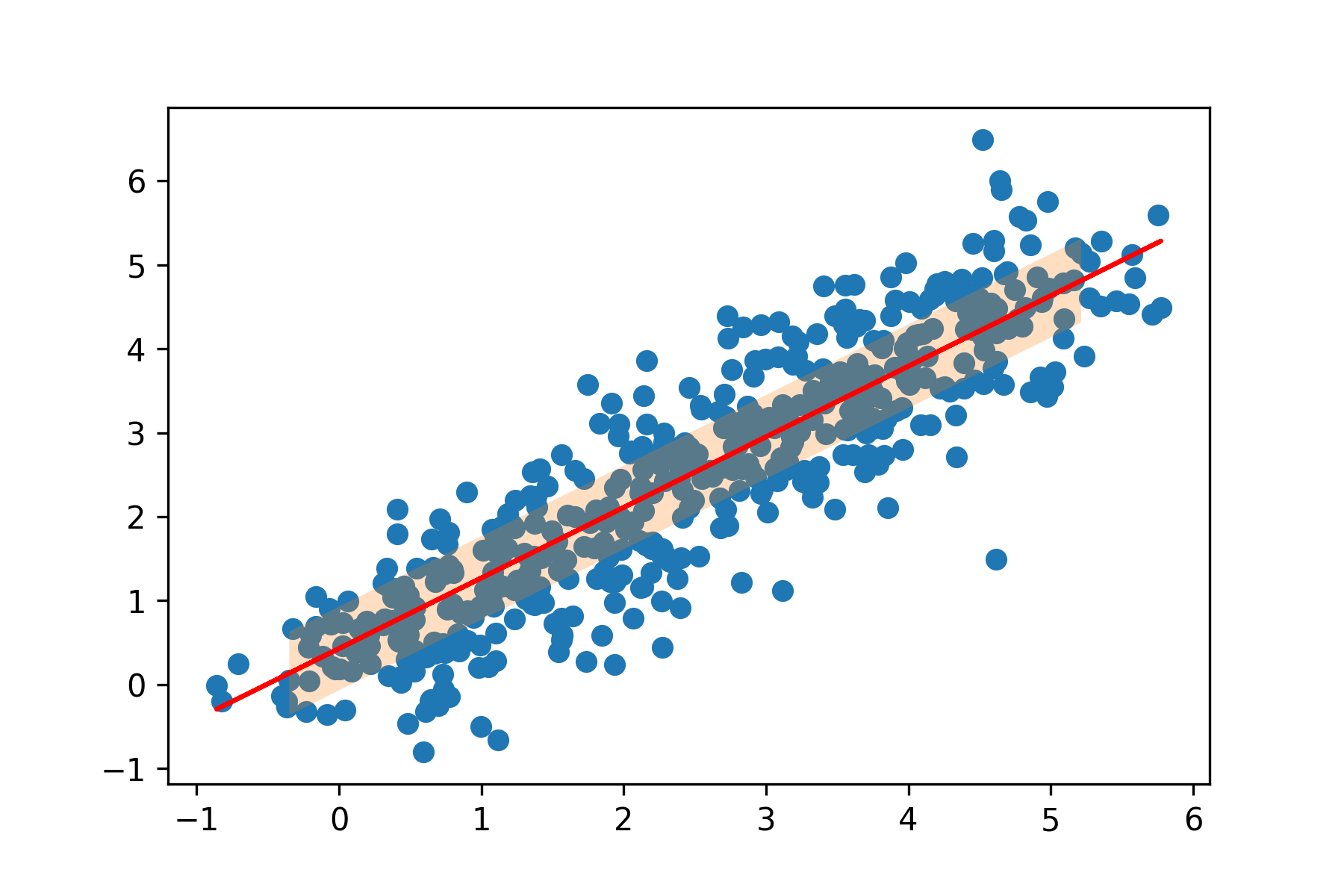} }}
    \caption{Left) Failed NP regression with a uniform noise model $Unif[-1,1]$ and 500 synthetic data points. The line represents the estimated line of best fit, while the tube represents the undertainty of the model. Since the model is misspecified, the likelihood (eq \ref{unif_lik}) is zero and there is no learning signal for parameter updates. Right) Successful WNP regression with an identical experimental setup. In this case, the model does not rely on likelihood, instead it makes use of the signal from the sliced Wasserstein distance which persists even in the misspecified case.}
    \label{fig:uniform}
\end{figure}

However, consider the case where the noise model is Uniform (e.g., $Unif[-1,1]$). There are now settings for $\beta,b$ under which the data $x$ would have exactly zero likelihood (i.e,, any time a single data point falls outside the "tube" of noise).  In the worst case, there may be \emph{no} setting of the parameters that provides non-zero likelihood.  Furthermore, because of the uniform nature of the noise likelihoods, there are no useful gradients.
\begin{align}
    L(\beta, b) = \Pi_{i \in I} p(y_i | x_i, \beta, b) \\
    = \Pi_{i \in I} 0.5 \cdot \mathbb{1}(|y_i - (\beta x + b)| < 1)
    \label{unif_lik}
\end{align}
In these cases, any model based on likelihoods, including NPs, would fail to learn.  This is illustrated in fig ~\ref{fig:uniform}.
%To be more precise, in the regression case with the aforementioned uniform noise model we would have that (eq \ref{unif_lik}) would be zero if any of the data lie outside the uniform tube.

However, even if the data likelihood is degenerate, there is still a well-defined notion of the optimal transport cost between the observed data and the distribution implied by the model, no matter what setting of parameters is used.  Since the Wasserstein distance is defined independent of likelihood, the WNP model still finds the proper fit of parameters given the data as seen in (fig \ref{fig:uniform}). This experiment could be analogous to the real world case where our prior knowledge of the data is fundamentally flawed. This flaw implies that our model is incorrect and misspecified. For this experiment, we generate synthetic data from $y = 1 \cdot x + 0 + \epsilon$ where $\epsilon \sim \mathcal{N}(0,0.5)$. The sliced Wasserstein distance is calculated as described in section \ref{slicedwasssect} with $N = 50$. We let $h$ and $g$ be two-layer fully connected Neural Networks and our $r_C$ vector is of size $32$ and we use the Adam \cite{KingmaB14} optimizer with defaults.

\subsection{Intractable Likelihoods - The Quantile "$g$-and-$\kappa$" Distribution}

We now present a second example, with a distribution that is easy to sample from, but for which the likelihood cannot be tractably computed.  The $g$-and-$\kappa$ distribution \cite{tukey1977, jorge1984some} is defined for $r \in (0,1)$ and parameters $\theta = (a,b,g,\kappa) \in [0,10]^4$ as:
\begin{equation}
    a + b \left(1 + 0.8 \frac{1 - \text{exp}(-gz(r))}{1 + \text{exp}(-gz(r))}\right)(1 + z(r)^2)^{\kappa}z(r)
\end{equation}
Where $z(r)$ refers to the $r$-th quantile of a standard Normal distribution $\mathcal{N}(0,1)$. It is known that the density function and likelihood are analytically intractable, making it a perfect test for WNPs. This model is a standard benchmark for approximate Bayesian computation methods \cite{sisson2018handbook}. Despite the intractability, it is straightforward to sample i.i.d. variables from this distribution by simply plugging in standard Normals in place of $z(r)$ \cite{bernton2019approximate}. 

\begin{figure}
    \centering
    \includegraphics[width=0.45\textwidth]{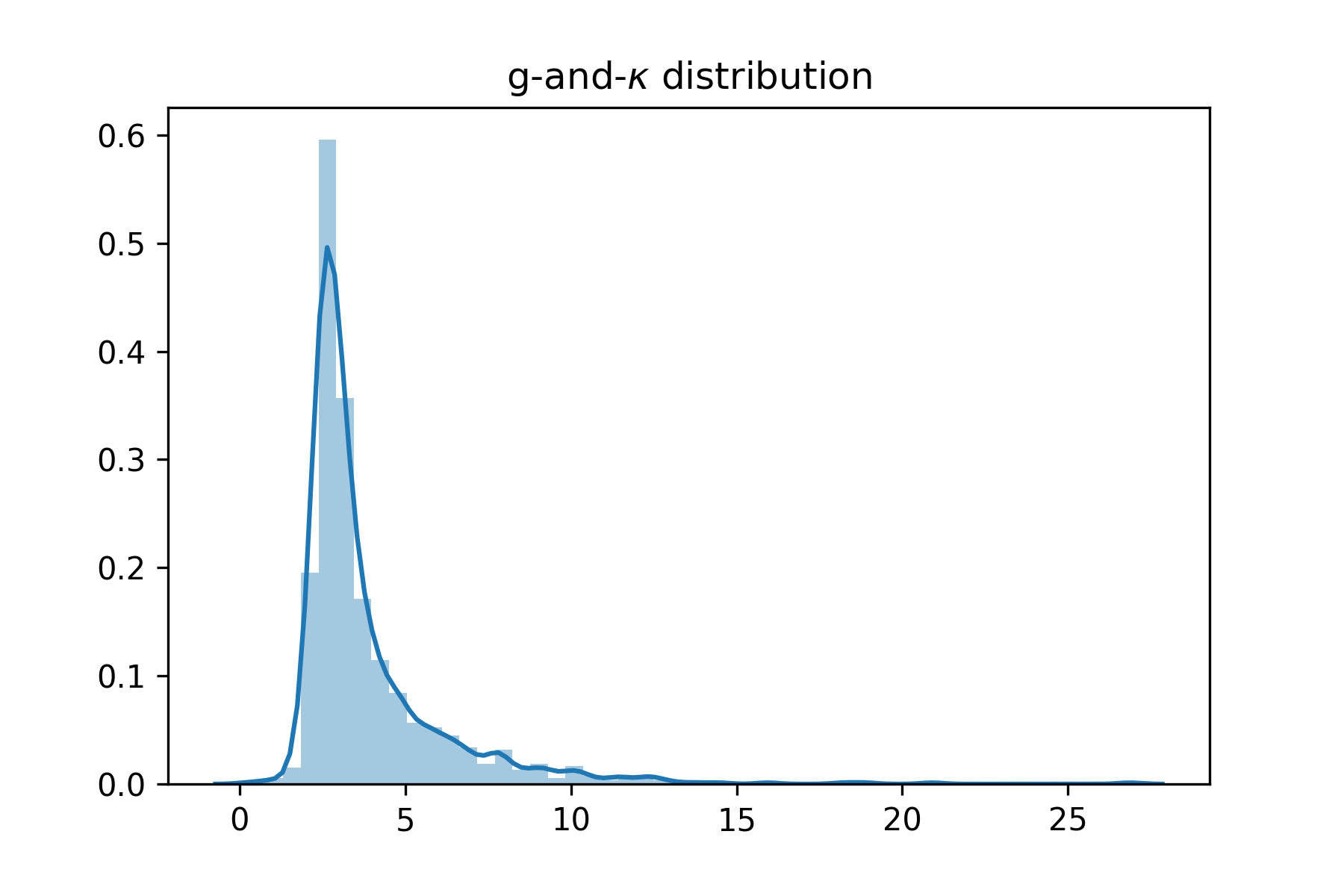}
    \includegraphics[width=0.45\textwidth]{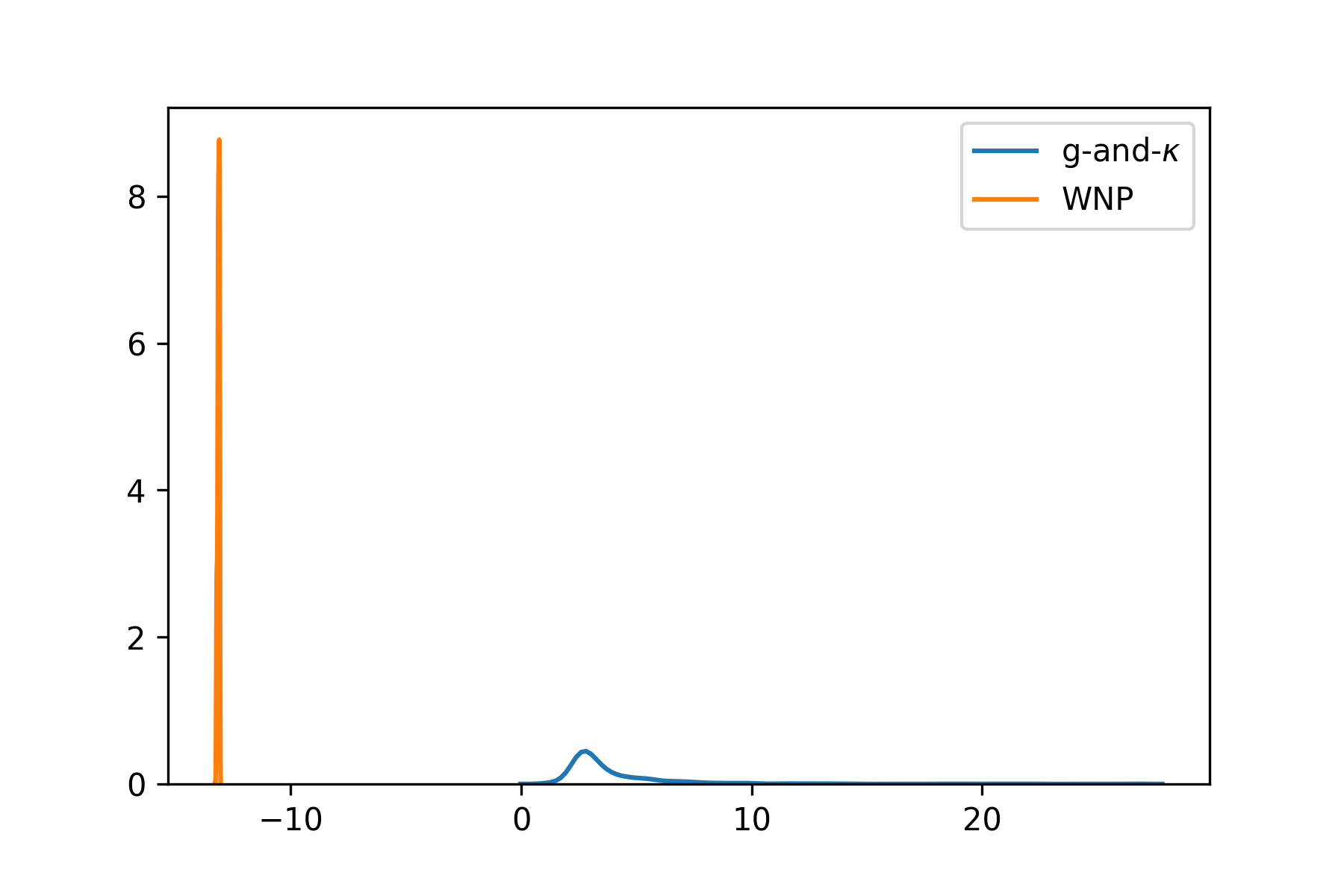}
    \includegraphics[width=0.45\textwidth]{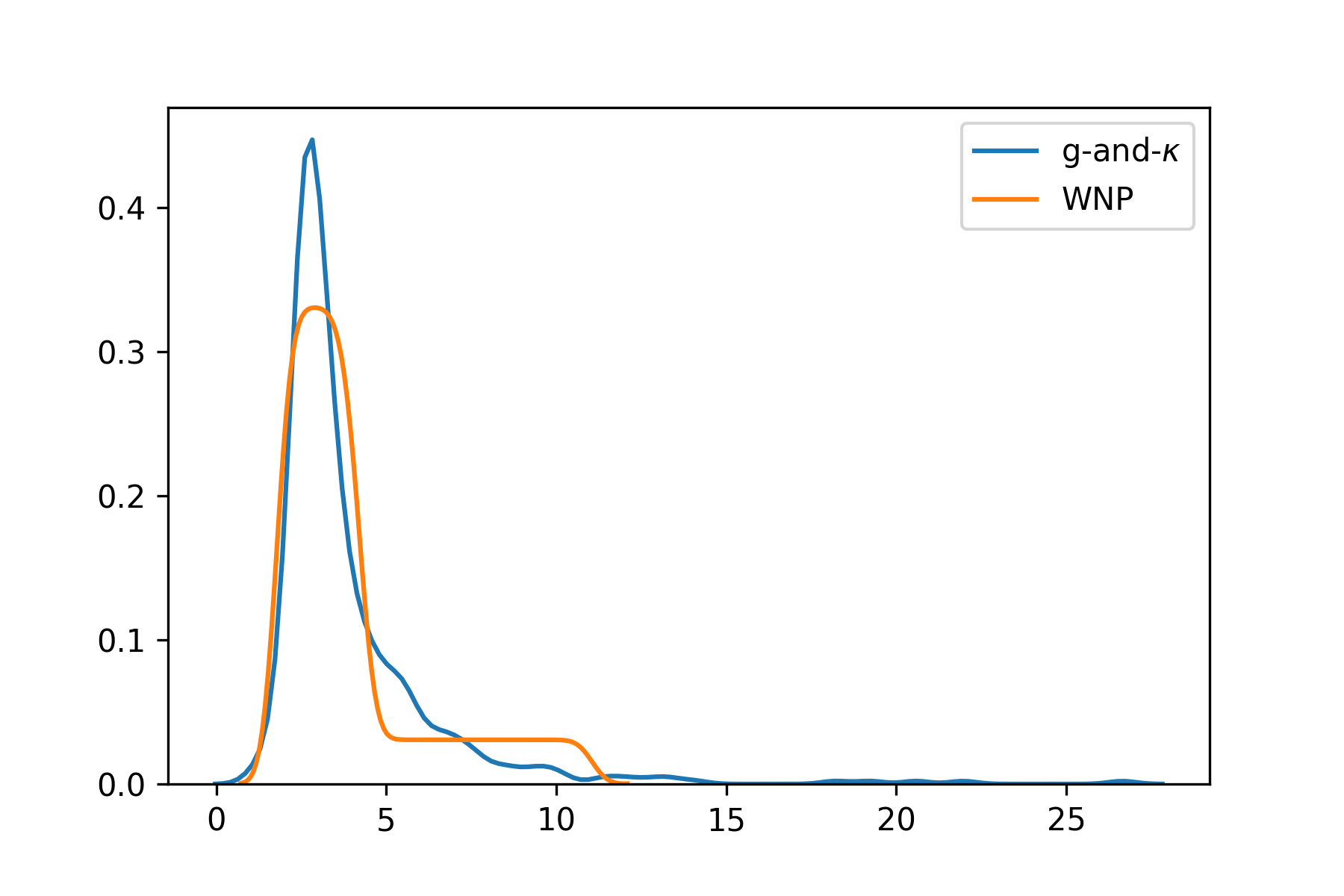}
    \includegraphics[width=0.45\textwidth]{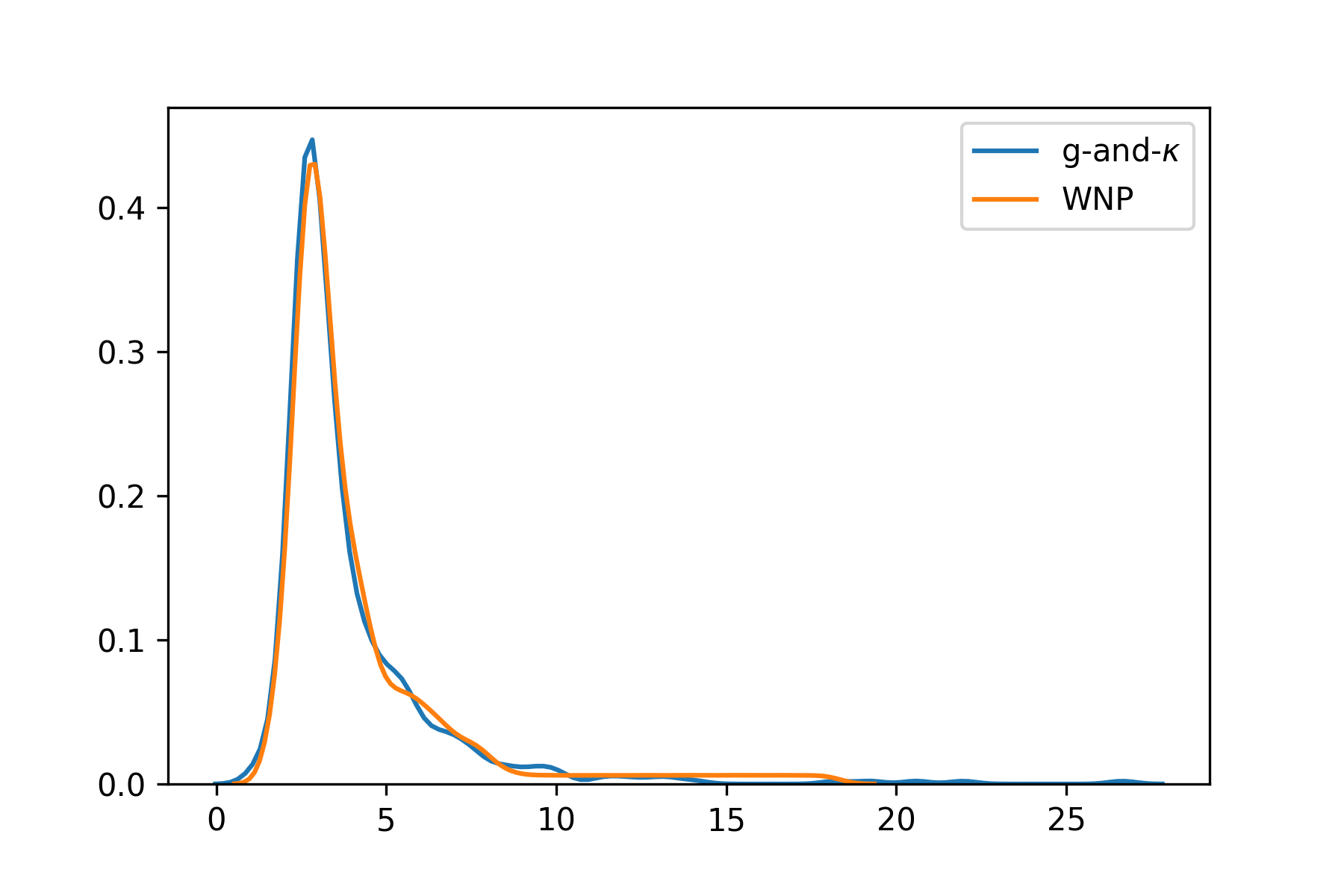}
    \caption{Top left: Initialization of learned distribution. It would be impossible to calculate the likelihood in this case for traditional NPs both because the model is misspecified and because the likelihood is computationally intractable. Top right: Beginning of the learning process. The output of the WNP is completely disjoint from the sampled $g$-and-$\kappa$ distribution. Bottom left: Part way through the learning process (approximately 500 steps) we see the WNP is quickly able to achieve a good fit. Bottom right: Final fit of $g$-and-$\kappa$ distribution where WNP learned to model the distribution without relying on the likelihood calculation}
    \label{fig:g_and_kappa}
\end{figure}

We follow the experimental set up in \cite{bernton2019approximate} and take $\theta = (3,1,2,0.5)$. Which produces a distribution seen in (fig \ref{fig:g_and_kappa}). Additionally, we let $h$ and $g$ be two-layer fully connected neural networks with slightly more parameters than in the uniform noise model experiment and with the same size $r_C$ vector ($32$ dimensional). In this case, we use a cyclic learning rate with the base learning rate $1e^{-3}$ and the max learning rate $1e^{-2}$. We found that in some cases, our model would output a shifted empirical distribution, and cycling the learning rate would finish the optimization process well. We hypothesize this as a function of the $g$-and-$\kappa$ topology and leave it as an area for future exploration as it is beyond the scope of this work.

% \begin{figure}[h]
% \centering

% \caption{Example of the $g$-and-$\kappa$ distribution with 1000 draws fit with KDE}
% \label{fig:g_and_kappa}
% \end{figure}

Due to the fact that the likelihood is intractable, if one wished to use this distribution as the model output of a Neural Process, it would be as impossible to train as in the uniform noise model regression task. This is due to the fact, as mentioned above, that standard Neural Processes rely on maximum likelihood as a learning signal. Since Wasserstein Neural Processes rely solely on the divergence between two empirical distributions, they can learn a conditional $g$-and-$\kappa$ distribution. 

\subsection{CelebA tiles as super pixels}

We now present our final experiment on high-dimensional image data. 
The purpose of this experiment is to illustrate the ability of WNP to model higher dimensional distributions where the likelihood might be difficult to calculate, or potentially meaningless. The experiment is run on the CelebA dataset. 

To distinguish between a large class of prior work and our contribution, we highlight a distinction between common candidate objective functions and the true data distribution. Many frameworks for image prediction, including standard Neural Processes implicitly, involve \emph{reconstruction-based} loss functions that are usually defined as $\lVert y - \hat{y}\rVert$.  These loss functions compare a target $y$ with the model's output $\hat{y}$ using pixel-wise MSE -- but this is known to perform poorly in many cases, including translations and brightness variations. More generally, it completely ignores valuable higher-order information in images: pixel intensities are neither independent nor do they share the same variance.  Similarly, it is well-known that images are typically supported on low-dimensional manifolds.

Put simply, pixel-wise losses imply false assumptions and results in a different objective than the one we would truly like to minimize. These false assumptions help explain why (for example) generative image models such as VAEs (based on reconstruction error) result in blurry reconstructions, while models such as Wasserstein GANs (based on Wasserstein divergences) result in much sharper reconstructions.

Neural Processes use Gaussian distributions over individual pixels.  Predictably, this results in blurry predictions, seen in \cite{kim2018attentive, NP:follow, CNP:baby}.  Analogously to the difference between GANs and VAEs, if we extend these experiments slightly, where instead of using pixel values as context points, we take large tiles of the image as context points, we might expect that we would see sharper reconstructions from WNPs as compared to NPs.

We test this by treating image completion as a high-dimensional regression task. 32x32 images are broken into 4x4 tiles; these tiles are then randomly subsampled.  The $X$ inputs are the tile coordinates; the $Y$ outputs are the associated pixels.  Given a set of between 4-16 tiles, the WNP must predict the remainder of the pixels by predicting entire tiles.  The Neural Process is trained on the same data, using the same neural network; the only difference is the loss function.

\begin{figure}
    \centering
    \subfloat{{\includegraphics[scale=0.5]{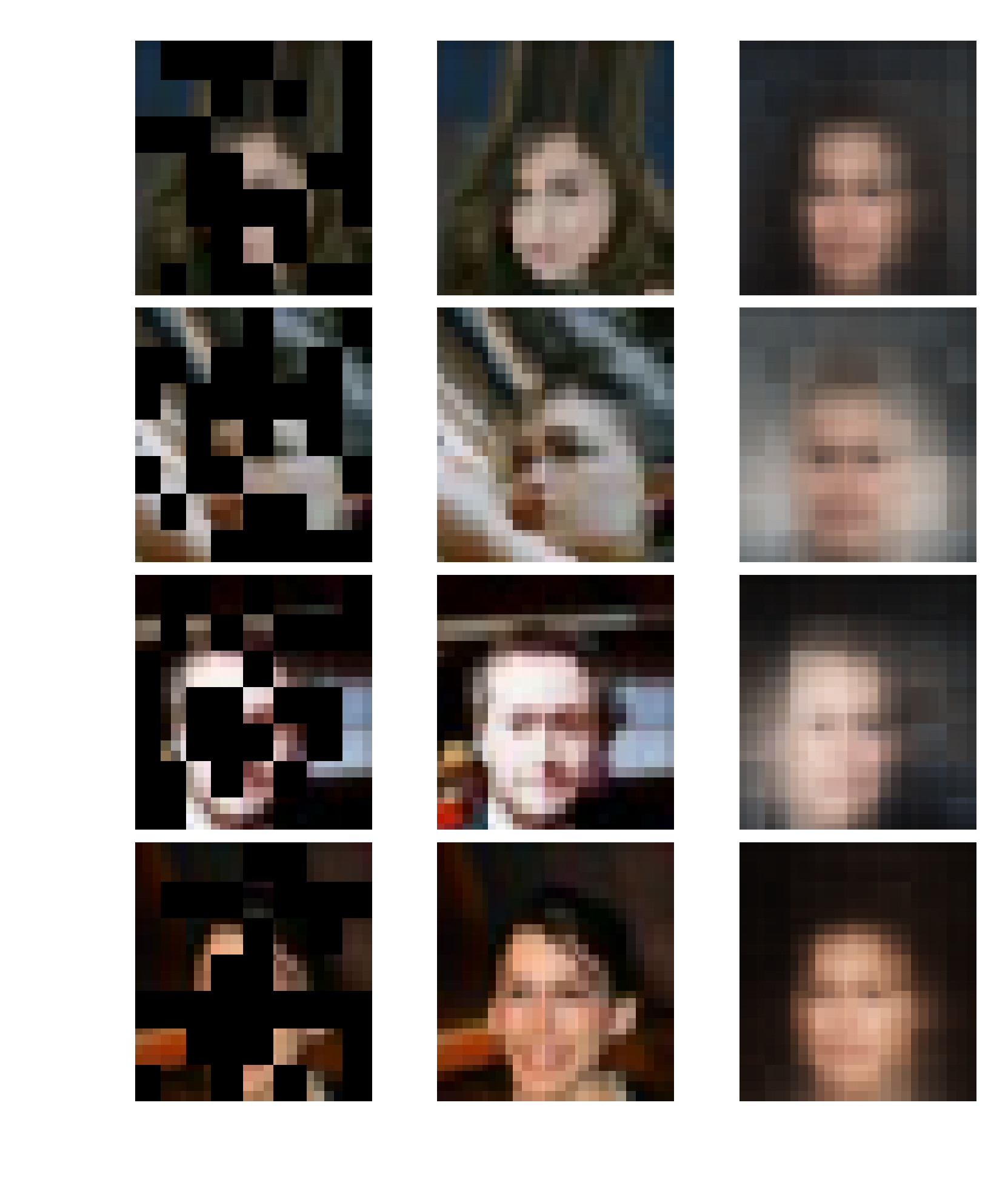}}}
    \subfloat{{\includegraphics[scale=0.5]{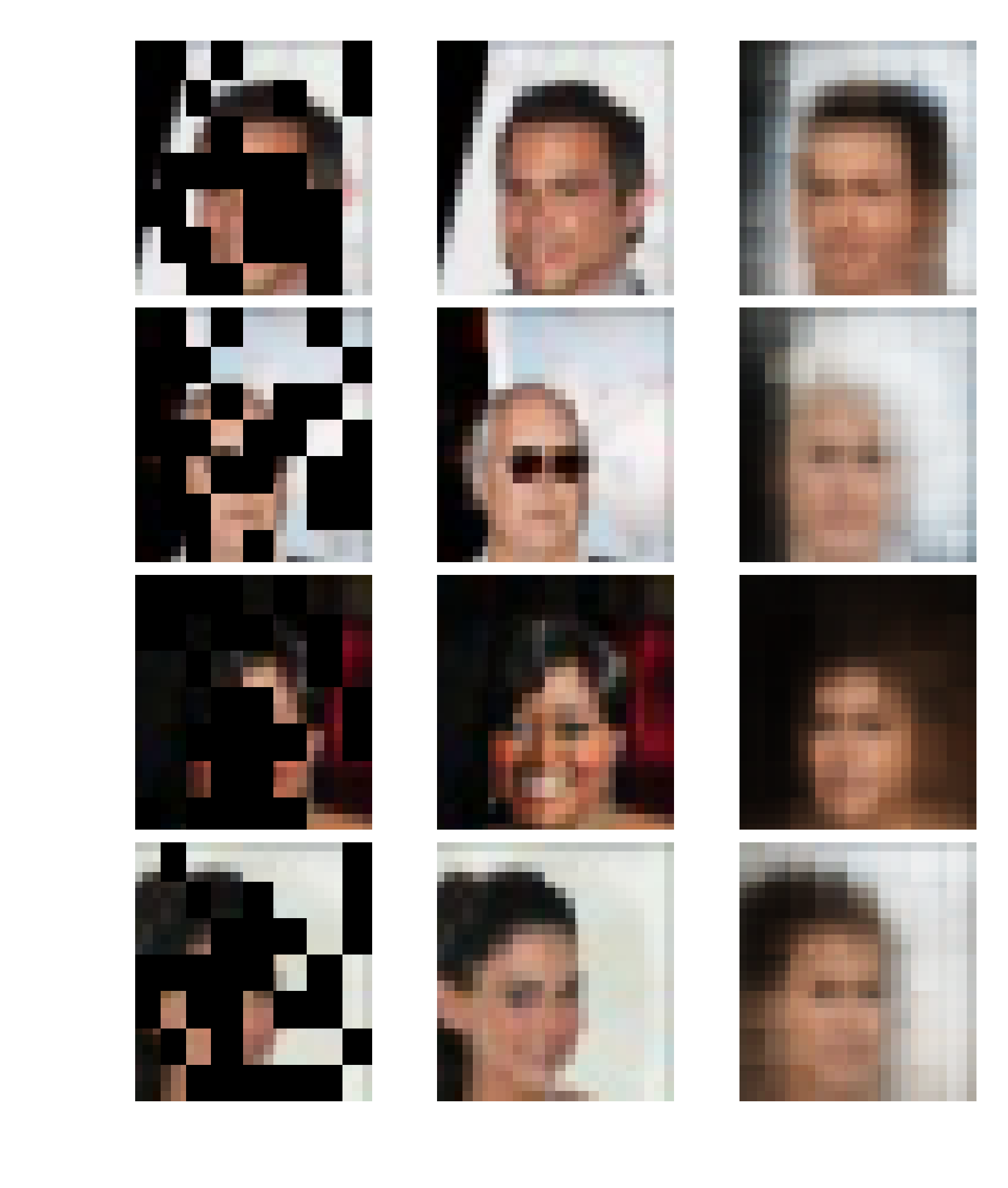}}}
    \caption{On the left: results from Neural Processes. The left-most column shows the sparse observations; the middle column shows ground truth; the right column shows the NP reconstruction. On the right: results for WNPs.  See text for discussion.}
    \label{fig:celeb-a}
\end{figure}

\subsection{Discussion}

The results of our Celeb-A experiment are shown in (fig ~\ref{fig:celeb-a}).  Here, the results are what we expected: traditional Neural Processes fail to learn more than an extremely blurry output when using tiles as context points instead of pixels. This is because the likelihood is now significantly less meaningful. However, in the case of WNP, the empirical sliced Wasserstein distance is still well defined and can learn a better conditional distribution. While the images are not as high quality as much recent GAN work, the purpose of this experiment was to illustrate that the WNP can better capture background, facial structure, color, than traditional NP trained with maximum likelihood. Qualitatively, we see that the images are slightly more clear and better capture the background and structural information of the image, given the context conditioning

\section{Conclusion}

In this work, we have explored a synthesis of Neural Processes and Optimal Transport. We demonstrated that there are desireable classes of functions which Neural Processes either struggle to learn, or cannot learn, but which can be learned by Wasserstein Neural Processes. These WNPs use Wasserstein distance in place of the intrinsic KL divergence used in maximum likelihood, and can be tractably implemented using (for example) sliced approximations. Maximum likelihood fails to learn in the case of misspecified models and likelihoods that either don't exist, or are computationally intractable.  As a concluding thought, we note that our technical strategy in developing this model was to replace a maximum-likelihood objective with a Wasserstein based objective.  This raises the intriguing question: can \emph{all} probabilistic models based on maximum-likelihood be reformulated in terms of Wasserstein divergence?  We leave this exciting direction for future work.

% \bibliographystyle{unsrt}
% \bibliography{ref}
\end{document}